\documentclass[10pt,twocolumn]{article}

\usepackage[a4paper,margin=2cm,columnsep=0.65cm]{geometry}
\usepackage[T1]{fontenc}
\usepackage[utf8]{inputenc}
\usepackage{booktabs}
\usepackage{float}
\usepackage{placeins}
\usepackage{graphicx}
\usepackage{amsmath}
\usepackage{multirow}
\usepackage{xcolor}
\usepackage[expansion=false]{microtype}  % expansion off: fails on bitmap-font TeX installs
\usepackage[natbibapa]{apacite}
\shortcites{rokuss2025dac,isensee2021nnunet,saha2021duke,garrucho2025mamamia,daniels2024ambl,comstock2023ea1141,wald2025crosstechnique,isensee2025nninteractive,huang2014qinbreast,huang2014qinchallenge,geissler2026nipple}
\defcitealias{sagittalLandmarks2026}{MRI Sagittal}
\usepackage[colorlinks=true,allcolors=blue]{hyperref}
\usepackage{authblk}

\title{Divide First: A Compact 3D Model for Per-Breast Region and Nipple
Landmark Segmentation in Breast DCE-MRI}

\author[1]{Hassan Alhajj}
\author[1]{Mickael Tardy}
\affil[1]{Hera-MI, SAS, Nantes, France}
\date{}

\begin{document}
\maketitle

\begin{abstract}
Together with breast-region segmentation, nipple position is essential for any breast MRI AI product. We present a model and inference code for joint breast-region and nipple segmentation in DCE-MRI. We trained deep learning models
on a nipple-labelled MRI dataset derived from the BreastDivider dataset and
compared them under two different setups: bilateral and per-breast.
The divide-first strategy processes each breast separately and restores
laterality geometrically. The compact per-breast
model achieves a sensitivity of 96.1\% with 0.104
false-positive nipple components per bilateral volume (FP).
We release the model, inference code and BreastDivider nipple prediction masks to
support reuse in breast MRI analysis. This release represents an initial
effort to make AI models for potential production workflows easier to
reproduce and accessible to the wider community. Evaluation on 65 unseen
sagittal cases is presented in Appendix~\ref{sec:sagittal}.
\end{abstract}

% ===========================================================================
\section{Introduction}
% ===========================================================================

Automated breast MRI analysis involves several processing steps. A production pipeline
typically combines region-of-interest extraction, lesion detection,
motion quality control and structured reporting. Errors in the region-of-interest step can affect
subsequent analyses. In this study, we focus on two anatomical outputs:

\begin{itemize}
  \item The \textbf{breast region}, which tells the pipeline which
        voxels belong to the left breast and which to the right, so that
        findings can be attributed to a side, bilateral exams can be split
        into independent halves, and post-mastectomy exams can be handled
        without assuming that both breasts are present.
  \item The \textbf{nipple landmark}, which is the conventional origin for
        clinical localisation: clock-face position and distance from the nipple are used to report the location of a finding~\citep{acr2013birads}. Nipple position provides an anatomical reference
        for matching findings across imaging modalities.
\end{itemize}

The ``Divide and Conquer'' (DAC) work introduced
\emph{BreastDivider}~\citep{rokuss2025dac}: the first large-scale public breast MRI dataset with explicit left and right breast
segmentation labels. Its first version comprises 13,752 3D scans from
Duke-Breast-\allowbreak Cancer-MRI~\citep{saha2021duke}, MAMA-MIA~\citep{garrucho2025mamamia},
Advanced-MRI-Breast-Lesions~\citep{daniels2024ambl} and
EA1141~\citep{comstock2023ea1141}, converted to NIfTI and oriented to RAS (i.e. 3D coordinate system where the x, y, and z axes point toward the patient's right, anterior, and superior respectively). The release also includes an nnU-Net~\citep{isensee2021nnunet} model trained for left and right breast
segmentation, with a reported mean Dice of 0.9908 in five-fold cross-validation. The model weights are released under CC BY-NC-SA 4.0.

In this work, we aim to provide reusable models and code for breast MRI
analysis. This analysis mainly consists of identifying the breast regions and nipple landmarks in a DCE-MRI. Our contribution
is twofold:

\begin{enumerate}
  \item We publicly release nipple masks predicted across the
        BreastDivider dataset through
        DAC-nipple-predictions~\citep{dacNipplePredictions2026}.
        These masks add nipple predictions to breast-region information. They
        are model outputs, distinct from the manually annotated, reviewed and propagated
        labels used to train and evaluate our nipple model
        (Section~\ref{sec:dataset}), and they include the volumes used for training. 
    \item We release a model and inference code for breast MRI analysis,
        combining breast-region segmentation and nipple localisation. The
        selected model processes each breast separately and restores
        laterality geometrically
        (Sections~\ref{sec:method}-\ref{sec:results}). The inference code
        also includes motion analysis between DCE phases, which is not
        evaluated in this paper.
\end{enumerate}

% ===========================================================================
\section{Related work}
% ===========================================================================

\paragraph{Breast region segmentation.}
DAC introduced a large, consistently labelled dataset for benchmarking breast-region segmentation. Its labels were bootstrapped from 100 Duke
whole-breast masks split at the sagittal plane through their centre of mass, then extended through an active-learning loop with
cross-sequence pseudo-labels, following the annotation transfer of ~\citep{wald2025crosstechnique}. All cases
were visually reviewed by a team of three experts and refined with
nnInteractive~\citep{isensee2025nninteractive} where needed.

\paragraph{Nipple localisation.}
Earlier automated methods on breast MRI locate the nipple with geometric rules,
such as a protrusion above the fitted breast contour~\citep{gwo2014nipple} or
the most anterior point of the body, used to compute the 3D tumour-to-nipple
distance for surgical planning~\citep{giannini2017nac,dalonzo2019nsm}. These
studies evaluate nipple detection sensitivity and centroid or Euclidean distances. Such rules fail on inverted nipples~\citep{dalonzo2019nsm}. Initial
learned approaches exist, such as active learning for nipple segmentation in
breast MRI~\citep{geissler2026nipple}, but to the best of our knowledge no
large-scale public breast MRI nipple dataset exists, at the scale of
BreastDivider or otherwise.

% ===========================================================================
\section{Dataset}
\label{sec:dataset}
% ===========================================================================

Version 2 of BreastDivider~\citep{rokuss2025dac} contains 17,956 3D breast MRI volumes from seven source datasets, with left
and right breast-region masks but no nipple masks.

We added nipple masks to 8,054 BreastDivider-derived 3D MRI volumes from 1,319 patients. These volumes span six cohorts: EA1141,
Advanced-MRI-Breast-Lesions (AMBL), Duke-Breast-\allowbreak Cancer-MRI,
ODELIA~\citep{mullerfranzes2025odelia}, and I-SPY1/I-SPY2 as represented in MAMA-MIA. Each phase or sequence
from one patient counts as a distinct volume. Table~\ref{tab:dataset} gives the
cohort-level volume counts. I-SPY1 contributes few volumes because only 4 of its MAMA-MIA patients have bilateral MRI.

\begin{table*}[t]
\centering
\caption{Nipple-labelled dataset by source cohort. Each MRI volume produces
two breasts for the divide-first configuration; these are not independent
patients.}
\label{tab:dataset}
\begin{tabular}{lrrrr}
\toprule
Cohort & Train & Validation & Total volumes & Breasts \\
\midrule
AMBL        & 2,325 & 209 & 2,534 & 5,068 \\
Duke           & 2,053 & 59  & 2,112 & 4,224 \\
EA1141   & 1,482 & 224 & 1,706 & 3,412 \\
ODELIA                             & 1,151 & 121 & 1,272 & 2,544 \\
I-SPY2  & 383  & 42  & 425  & 850  \\
I-SPY1  & 5    & 0   & 5    & 10   \\
\midrule
\textbf{Total} & 7,399 & 655 & 8,054 & 16,108 \\
\bottomrule
\end{tabular}
\end{table*}

%The source collections differ in scanner, field strength, sequence and
%acquisition protocol; EA1141 alone spans 48 clinical sites. We select one
%3D volume at a time for model input, while preserving patient grouping across
%repeated volumes for the train--validation split.

We used the left and right breast-region masks supplied by BreastDivider without modification. They were produced by the DAC labelling process described above and are partly model-derived.
%Consequently, breast-region Dice measures agreement with these curated,
%partly model-derived labels. It is not an estimate of agreement with an
%independent, fully manual anatomical reference.

\subsection{Nipple landmark annotation}

We began with 350 manually drawn nipple masks from 320 patients.
The first expansion copied a reviewed seed mask to other phases of the same exam only
when their geometry matched. Subsequent iterations trained a 3D U-Net (described in Section~\ref{sec:architectures}) in all three rounds to generate
candidate masks. At each iteration, every
volume in the review pool was inspected at least once in a dedicated viewer
to assess the predicted mask and decide whether to accept it or select it
for correction in 3D Slicer. Reviewed masks were also propagated to compatible
phases of the same patient. Repeating this
annotation and review process over three iterations yielded nipple masks for
8,054 MRI volumes. A single annotator performed all annotation and review.
% Nipple labels are volumetric masks. 
%Evaluation uses connected-component
%centroids for localisation and records missed references separately from
%distance among matched components. The 655-volume validation split yields
%1,310 breast half-volumes and 1,300 reference nipple components; 15 halves
%have empty nipple masks. Empty masks remain in false-positive accounting.
%These aggregate counts do not by themselves identify the reason for an empty
%mask or imply exactly one component in every nonempty half. The separately
%published DAC nipple prediction masks~\citep{dacNipplePredictions2026}
%are inference outputs across DAC, not the annotation set used to train or
%evaluate this model.

% ===========================================================================
\section{Method}
\label{sec:method}
% ===========================================================================

All DCE-MRI volumes are reoriented to canonical RAS and resampled to a 1.5\,mm
isotropic grid, using third-order spline interpolation for images and, for labels, linear interpolation of each label channel followed by thresholding at 0.5. In the divide-first setup, each
volume is split at the midline of its field of view after reorientation and before
resampling, and the two halves are resampled independently. All experiments
use the same patient-disjoint partition: 7,399 training volumes from 1,201
patients and 655 validation volumes from 118 patients, with all
volumes of a patient assigned to the same partition.

\subsection{Input}
\label{sec:Input}

\paragraph{Bilateral.} The network processes the whole examination to segment
the breast tissue, assign each voxel to a side and predict nipple masks for
both breasts.

\paragraph{Divide-first.} The midline decision is made geometrically, outside
the network. The network processes one breast and segments its region and nipple. Laterality is assigned from the side of the split. The split is applied only to exams marked as bilateral, assuming a field of
view centred on the patient; exams marked as unilateral are processed whole in
a single pass, and this bilateral or unilateral setting is supplied with each
exam rather than detected from the image.

% Defined here, a page early: a two-column float appears at the earliest on
% the next page, which places it on the Results page.
\begin{table*}[t]
\centering
\caption{Model checkpoint selected for each configuration as the one with the fewest nipple false positives among the ones
with sensitivity~$\ge$~96\%. Sensitivity, FP and distance are defined in Section~\ref{sec:protocol}.
Brackets give 95\% bootstrap confidence intervals.}
\label{tab:main}
\footnotesize
\setlength{\tabcolsep}{2pt}
\begin{tabular}{llcccc}
\toprule
 & & Breast region & \multicolumn{3}{c}{Nipple landmarks} \\
\cmidrule(lr){3-3}\cmidrule(lr){4-6}
Model & Configuration & Dice & Sens (\%) & FP & Distance (mm) \\
\midrule
Compact & bilateral    & 0.9687 [0.9676, 0.9697] & 96.1 [95.0, 97.2] & 0.139 [0.110, 0.168] & 3.79 [3.10, 4.62] \\
nnU-Net & bilateral    & 0.9727 [0.9716, 0.9737] & 96.1 [94.8, 97.3] & 0.095 [0.069, 0.124] & 3.71 [3.06, 4.44] \\
Compact & divide-first & 0.9730 [0.9721, 0.9738] & 96.1 [95.0, 97.2] & 0.104 [0.075, 0.133] & 3.41 [2.85, 4.07] \\
nnU-Net & divide-first & 0.9703 [0.9693, 0.9713] & 96.0 [94.8, 97.3] & 0.136 [0.102, 0.171] & 3.35 [2.78, 4.00] \\
\bottomrule
\end{tabular}
\end{table*}

\subsection{Network architectures}
\label{sec:architectures}
We compare two networks: the nnU-Net architecture \citep{isensee2021nnunet} with the plan used in DAC, with 26.1 million parameters, and a 3D U-Net
with an EfficientNet-B0 encoder and 11.4 million parameters. The nnU-Net architecture is trained in our pipeline with the training procedure described below, not with the self-configuring nnU-Net framework. Each network
is trained jointly for breast-region and nipple segmentation using two output heads, one for each task. For a patch with $N$ voxels,
let $p_i$ be the predicted probability at voxel $i$ and $y_i$ its binary target. The binary
loss combines binary cross-entropy (BCE) and soft Dice:
\begin{align}
D(p,y) &= \frac{2\sum_i p_i y_i+1}{\sum_i p_i+\sum_i y_i+1}, \\
\mathcal{L}_{\mathrm{BCE}}(p,y)
&= -\frac{1}{N}\sum_i \bigl[y_i\log p_i \nonumber\\
&\qquad +(1-y_i)\log(1-p_i)\bigr], \\
\mathcal{L}_{\mathrm{bin}}(p,y)
&= \mathcal{L}_{\mathrm{BCE}}(p,y)+1-D(p,y).
\end{align}
For the bilateral
breast-region head, $q_{ic}$ is the softmax probability for class
$c\in\{0,1,2\}$ (background, left and right), and $t_{ic}$ is its one-hot
target. Its loss is
\begin{align}
\mathcal{L}_{\mathrm{CE}}(q,t)
&= -\frac{1}{N}\sum_i\sum_{c=0}^{2}t_{ic}\log q_{ic},\\
\mathcal{L}_{\mathrm{multi}}(q,t)
&=\mathcal{L}_{\mathrm{CE}}(q,t)\nonumber\\
&\quad+\frac{1}{|F|}\sum_{c\in F}\bigl[1-D(q_c,t_c)\bigr],
\end{align}
where $F=\{c\in\{1,2\}:\sum_i t_{ic}>0\}$ contains the foreground classes
present in the patch. The Dice term is zero when $F$ is empty; background
remains supervised by cross-entropy. The nipple head always uses
$\mathcal{L}_{\mathrm{bin}}$, while the breast-region head uses
$\mathcal{L}_{\mathrm{bin}}$ for per-breast inputs and
$\mathcal{L}_{\mathrm{multi}}$ for bilateral inputs. For patch $b$,
$\mathcal{L}_{\mathrm{breast}}^{(b)}$ denotes the loss for its
breast-region head: $\mathcal{L}_{\mathrm{bin}}$ for per-breast inputs or
$\mathcal{L}_{\mathrm{multi}}$ for bilateral inputs. Likewise,
$\mathcal{L}_{\mathrm{nipple}}^{(b)}$ denotes the binary loss
$\mathcal{L}_{\mathrm{bin}}$ for the nipple head. For a batch of $B$ patches,
the joint objective is
\begin{equation}
\mathcal{L}=\frac{1}{B}\sum_{b=1}^{B}
\left(\mathcal{L}_{\mathrm{breast}}^{(b)}
+\mathcal{L}_{\mathrm{nipple}}^{(b)}\right),
\end{equation}
with both task weights set to 1.0.

\subsection{Training}

Both networks use the same patient partition for multi-task training. The inputs are processed per configuration as described in section ~\ref{sec:Input}. Input
volumes are cropped to the non-zero image region and z-score normalised per volume. Inference uses sliding-window prediction with 128$^3$ patches, 50\% overlap and Gaussian weighting, followed by the same post-processing for all experiments: objects smaller
than 20\,mm$^3$ for nipples and 100\,mm$^3$ for breast regions are removed,
and holes smaller than 1,000 and 2,000\,mm$^3$, respectively, are filled.
Predicted masks are restored to the original image grid.

\label{sec:protocol}
All experiments are evaluated on the same validation set. A reference nipple
is detected when a predicted nipple component overlaps more than 10\% of its
voxels. Sensitivity is the mean of reference nipples detected per input (bilateral or per-breast volume), and
FP is the mean number of predicted nipple components per bilateral volume that
detect no reference nipple. For distance measurement, we use the
Euclidean distance (in mm) between the centres of the reference and
predicted nipple masks. For bilateral inputs, we pair nipples across the
whole volume. For divide-first inputs, we pair nipples separately within
each breast. In each input, we keep as many of the largest predicted components as there are reference nipples, and pair them one-to-one to minimise the sum of distances. We average the
distances first within each input and then across inputs with at least one
pair. Breast-region segmentation is assessed using Dice.
All metrics are reported with 95\% bootstrap confidence intervals.

% ===========================================================================
\section{Results}
\label{sec:results}

\subsection{Breast region}

Dice ranged from 0.9687 to 0.9730 (Table~\ref{tab:main}). The compact
divide-first model scored highest (0.9730), followed by bilateral nnU-Net
(0.9727). Differences in data partitions and protocols preclude a direct
comparison with DAC's mean Dice of 0.9908. Figures~\ref{fig:landmarks-detail}
and~\ref{fig:landmarks-gallery} compare predicted and reference masks in a
detailed case and across five cohorts, respectively.

% Span both columns with the figures, then resume two-column Results text.
\twocolumn[{%
\begin{minipage}{\textwidth}
\begin{figure}[H]
\centering
\includegraphics[width=0.95\textwidth]{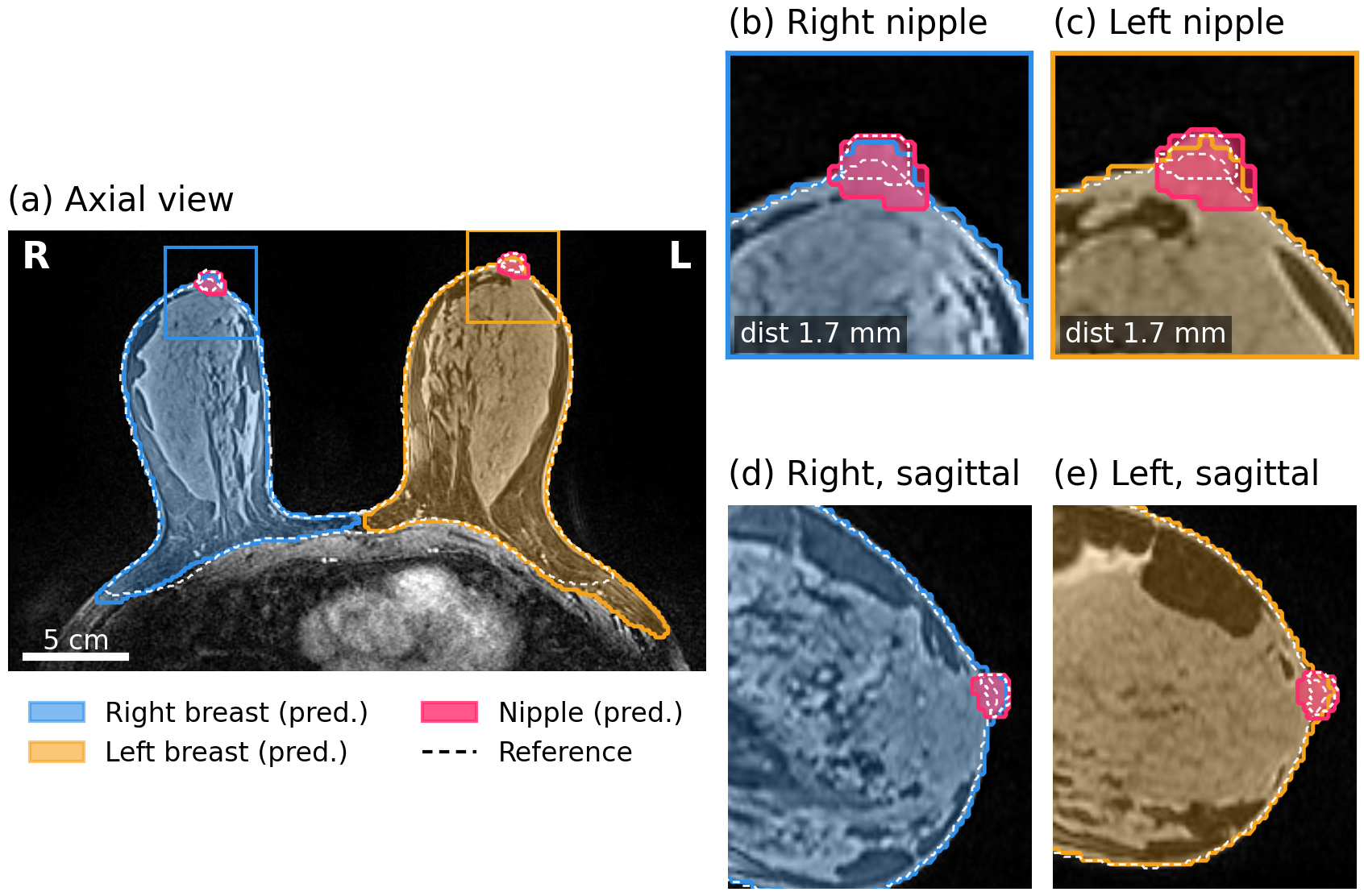}
\caption{Breast-region and nipple segmentation outputs. Predicted right and left breast regions are
shown in blue and orange, respectively, and predicted nipples in pink.
Dashed contours show the reference masks. Distances in (b, c) are measured
between the predicted and reference nipple centroids.}
\label{fig:landmarks-detail}
\end{figure}

\begin{figure}[H]
\centering
\includegraphics[width=\textwidth]{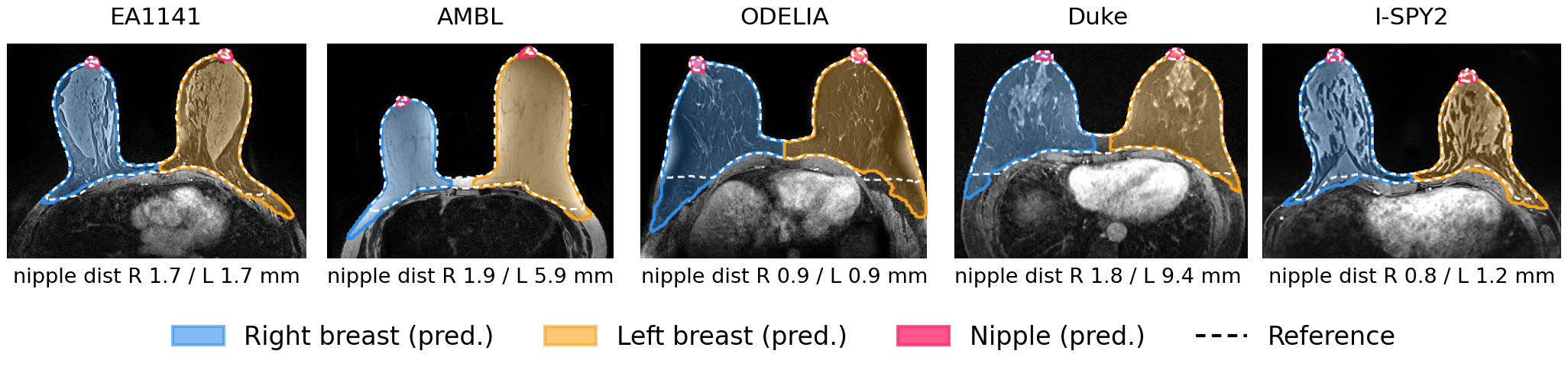}
\caption{Axial examples of breast-region and nipple segmentation from five
cohorts. Coloured overlays show the predictions, and dashed contours show
the reference masks. The right (R) and left (L) nipple centroid distances
are shown below each example.}
\label{fig:landmarks-gallery}
\end{figure}
\vspace{12pt}
\end{minipage}
}]

\subsection{Nipple landmarks}

All experiments detect 96.0\%-96.1\% of reference nipples (Table~\ref{tab:main}). FP
ranges from 0.095 for the bilateral nnU-Net architecture to 0.139 for the bilateral compact
network, with overlapping confidence intervals. The compact divide-first model has an FP of 0.104.

\newpage
Splitting reduces FP for the compact network (0.139 to 0.104) and increases it for the nnU-Net architecture
(0.095 to 0.136).
Mean nipple centroid distances ranged from 3.35 to 3.79\,mm across the four
configurations, with a mean of 3.41\,mm for the compact divide-first model.

\clearpage
\section{Conclusions}
\label{sec:conclusions}

We presented deep learning models for joint breast-region segmentation and nipple localisation. The results support the divide-first configuration with relatively simple 3D networks.
The compact divide-first model achieves a nipple sensitivity of 96.1\%, an FP
of 0.104, a mean centroid distance of 3.41\,mm for paired nipples, and a
breast-region Dice of 0.9730 (Table~\ref{tab:main}). We selected the compact divide-first model for release on Hugging Face. All
experiments reach the same sensitivity of about 96\%, and the selected model's FP of 0.104 is comparable to the lowest one (0.095, nnU-Net bilateral). It also has the highest breast-region Dice (0.9730) and uses fewer than half the
parameters of the nnU-Net architecture, which reduces memory use and inference cost. Processing each breast independently also accommodates unilateral inputs, and the sagittal evaluation provides evidence of transfer to unseen acquisition geometry (Appendix~\ref{sec:sagittal}, Table~\ref{tab:sagittal}), although detection and localisation errors remain. 

Breast-region Dice remains below the mean Dice of 0.9908 reported by DAC. One possible explanation is the joint optimisation of the breast-region and nipple heads: segmenting the much smaller nipple structures may introduce competing demands on the shared representation and reduce performance on the breast-region task. However, this effect was not isolated in a single-task comparison, and differences in the data partitions and evaluation protocols may also contribute to the observed gap.

The released model, inference code and nipple prediction masks provide reusable resources for further breast MRI analysis. These findings support the potential integration of the approach into clinical MRI analysis workflows, subject to further evaluation in the intended clinical setting.

% ===========================================================================
\section*{Data and code availability}
The nipple annotations used for training and validation are not released. The model weights, the inference project and the model card are available in the project repository~\citep{heramiMriAnalysisPaper2026}. The weights and the supplied inference code
are released under CC BY-NC-SA 4.0. The sagittal QIN cohort with its annotations is publicly available~\citepalias{sagittalLandmarks2026}
under CC BY 3.0. The upstream BreastDivider dataset and model are available from the original release~\citep{rokuss2025dac}; the model weights are licensed under CC BY-NC-SA 4.0.

\bibliographystyle{apacite}
\bibliography{references}

\begin{thebibliography}{}

\bibitem [\protect \citeauthoryear {%
{American College of Radiology}%
}{%
{American College of Radiology}%
}{%
{\protect \APACyear {2013}}%
}]{%
acr2013birads}
\APACinsertmetastar {%
acr2013birads}%
\begin{APACrefauthors}%
{American College of Radiology}.%
\end{APACrefauthors}%
\unskip\
\newblock
\APACrefYear{2013}.
\newblock
\APACrefbtitle {{ACR BI-RADS} Atlas: Breast Imaging Reporting and Data System}
  {{ACR BI-RADS} atlas: Breast imaging reporting and data system}\
  (\PrintOrdinal{5th}\ \BEd).
\newblock
\APACaddressPublisher{Reston, VA}{American College of Radiology}.
\PrintBackRefs{\CurrentBib}

\bibitem [\protect \citeauthoryear {%
Comstock%
\ \protect \BOthers {.}}{%
Comstock%
\ \protect \BOthers {.}}{%
{\protect \APACyear {2023}}%
}]{%
comstock2023ea1141}
\APACinsertmetastar {%
comstock2023ea1141}%
\begin{APACrefauthors}%
Comstock, C.%
, Gatsonis, C.%
, Newstead, G.%
, Snyder, B.%
, Gareen, I.%
, Bergin, J.%
\BDBL {}others%
\end{APACrefauthors}%
\unskip\
\newblock
\APACrefYearMonthDay{2023}{}{}.
\newblock
\APACrefbtitle {Abbreviated breast MRI and digital tomosynthesis mammography in
  screening women with dense breasts (EA1141).} {Abbreviated breast mri and
  digital tomosynthesis mammography in screening women with dense breasts
  (ea1141).}
\newblock
\APAChowpublished {The Cancer Imaging Archive, Little Rock, AR, USA}.
\PrintBackRefs{\CurrentBib}

\bibitem [\protect \citeauthoryear {%
D'Alonzo%
\ \protect \BOthers {.}}{%
D'Alonzo%
\ \protect \BOthers {.}}{%
{\protect \APACyear {2019}}%
}]{%
dalonzo2019nsm}
\APACinsertmetastar {%
dalonzo2019nsm}%
\begin{APACrefauthors}%
D'Alonzo, M.%
, Martincich, L.%
, Fenoglio, A.%
, Giannini, V.%
, Cellini, L.%
, Liberale, V.%
\BCBL {}\ \BBA {} Biglia, N.%
\end{APACrefauthors}%
\unskip\
\newblock
\APACrefYearMonthDay{2019}{}{}.
\newblock
{\BBOQ}\APACrefatitle {Nipple-sparing mastectomy: external validation of a
  three-dimensional automated method to predict nipple occult tumour
  involvement on preoperative breast {MRI}} {Nipple-sparing mastectomy:
  external validation of a three-dimensional automated method to predict nipple
  occult tumour involvement on preoperative breast {MRI}}.{\BBCQ}
\newblock
\APACjournalVolNumPages{European Radiology Experimental}{3}{}{31}.
\newblock
\begin{APACrefDOI} \doi{10.1186/s41747-019-0108-3} \end{APACrefDOI}
\PrintBackRefs{\CurrentBib}

\bibitem [\protect \citeauthoryear {%
Daniels%
, Last%
, Cohen%
, Mardor%
\BCBL {}\ \BBA {} Sklair-Levy%
}{%
Daniels%
\ \protect \BOthers {.}}{%
{\protect \APACyear {2024}}%
}]{%
daniels2024ambl}
\APACinsertmetastar {%
daniels2024ambl}%
\begin{APACrefauthors}%
Daniels, D.%
, Last, D.%
, Cohen, K.%
, Mardor, Y.%
\BCBL {}\ \BBA {} Sklair-Levy, M.%
\end{APACrefauthors}%
\unskip\
\newblock
\APACrefYearMonthDay{2024}{}{}.
\newblock
\APACrefbtitle {Standard and delayed contrast-enhanced MRI of malignant and
  benign breast lesions with histological and clinical supporting data
  (Advanced-MRI-Breast-Lesions), version 2.} {Standard and delayed
  contrast-enhanced mri of malignant and benign breast lesions with
  histological and clinical supporting data (advanced-mri-breast-lesions),
  version 2.}
\newblock
\APAChowpublished {The Cancer Imaging Archive}.
\newblock
\begin{APACrefDOI} \doi{10.7937/C7X1-YN57} \end{APACrefDOI}
\PrintBackRefs{\CurrentBib}

\bibitem [\protect \citeauthoryear {%
Garrucho%
\ \protect \BOthers {.}}{%
Garrucho%
\ \protect \BOthers {.}}{%
{\protect \APACyear {2025}}%
}]{%
garrucho2025mamamia}
\APACinsertmetastar {%
garrucho2025mamamia}%
\begin{APACrefauthors}%
Garrucho, L.%
, Kushibar, K.%
, Reidel, C\BHBI A.%
, Joshi, S.%
, Osuala, R.%
, Tsirikoglou, A.%
\BDBL {}others%
\end{APACrefauthors}%
\unskip\
\newblock
\APACrefYearMonthDay{2025}{}{}.
\newblock
{\BBOQ}\APACrefatitle {A large-scale multicenter breast cancer DCE-MRI
  benchmark dataset with expert segmentations} {A large-scale multicenter
  breast cancer dce-mri benchmark dataset with expert segmentations}.{\BBCQ}
\newblock
\APACjournalVolNumPages{Scientific Data}{12}{1}{453}.
\PrintBackRefs{\CurrentBib}

\bibitem [\protect \citeauthoryear {%
Geissler%
\ \protect \BOthers {.}}{%
Geissler%
\ \protect \BOthers {.}}{%
{\protect \APACyear {2026}}%
}]{%
geissler2026nipple}
\APACinsertmetastar {%
geissler2026nipple}%
\begin{APACrefauthors}%
Geissler, K.%
, Wenzel, M.%
, Diekmann, S.%
, Grimm, R.%
, von Busch, H.%
, Vik, T.%
\BCBL {}\ \BBA {} Meine, H.%
\end{APACrefauthors}%
\unskip\
\newblock
\APACrefYearMonthDay{2026}{}{}.
\newblock
{\BBOQ}\APACrefatitle {Applying Active Learning to Nipple Segmentation in
  Breast {MRI}} {Applying active learning to nipple segmentation in breast
  {MRI}}.{\BBCQ}
\newblock
\BIn{} \APACrefbtitle {Bildverarbeitung f{\"u}r die Medizin 2026}
  {Bildverarbeitung f{\"u}r die medizin 2026}\ (\BPGS\ 277--284).
\newblock
\begin{APACrefDOI} \doi{10.1007/978-3-658-51100-5_56} \end{APACrefDOI}
\PrintBackRefs{\CurrentBib}

\bibitem [\protect \citeauthoryear {%
Giannini%
\ \protect \BOthers {.}}{%
Giannini%
\ \protect \BOthers {.}}{%
{\protect \APACyear {2017}}%
}]{%
giannini2017nac}
\APACinsertmetastar {%
giannini2017nac}%
\begin{APACrefauthors}%
Giannini, V.%
, Bianchi, V.%
, Carabalona, S.%
, Mazzetti, S.%
, Maggiorotto, F.%
, Kubatzki, F.%
\BDBL {}Martincich, L.%
\end{APACrefauthors}%
\unskip\
\newblock
\APACrefYearMonthDay{2017}{}{}.
\newblock
{\BBOQ}\APACrefatitle {{MRI} to predict nipple-areola complex ({NAC})
  involvement: An automatic method to compute the {3D} distance between the
  {NAC} and tumor} {{MRI} to predict nipple-areola complex ({NAC}) involvement:
  An automatic method to compute the {3D} distance between the {NAC} and
  tumor}.{\BBCQ}
\newblock
\APACjournalVolNumPages{Journal of Surgical Oncology}{116}{8}{1069--1078}.
\newblock
\begin{APACrefDOI} \doi{10.1002/jso.24788} \end{APACrefDOI}
\PrintBackRefs{\CurrentBib}

\bibitem [\protect \citeauthoryear {%
Gwo%
, Gwo%
, Wei%
\BCBL {}\ \BBA {} Huang%
}{%
Gwo%
\ \protect \BOthers {.}}{%
{\protect \APACyear {2014}}%
}]{%
gwo2014nipple}
\APACinsertmetastar {%
gwo2014nipple}%
\begin{APACrefauthors}%
Gwo, C\BHBI Y.%
, Gwo, A.%
, Wei, C\BHBI H.%
\BCBL {}\ \BBA {} Huang, P\BPBI J.%
\end{APACrefauthors}%
\unskip\
\newblock
\APACrefYearMonthDay{2014}{}{}.
\newblock
{\BBOQ}\APACrefatitle {Identification of breast contour for nipple segmentation
  in breast magnetic resonance images} {Identification of breast contour for
  nipple segmentation in breast magnetic resonance images}.{\BBCQ}
\newblock
\APACjournalVolNumPages{Medical Physics}{41}{2}{022304}.
\newblock
\begin{APACrefDOI} \doi{10.1118/1.4861709} \end{APACrefDOI}
\PrintBackRefs{\CurrentBib}

\bibitem [\protect \citeauthoryear {%
{Hera-MI}%
}{%
{Hera-MI}%
}{%
{\protect \APACyear {2026}}%
{\protect \APACexlab {{\protect \BCnt {1}}}}}]{%
dacNipplePredictions2026}
\APACinsertmetastar {%
dacNipplePredictions2026}%
\begin{APACrefauthors}%
{Hera-MI}.%
\end{APACrefauthors}%
\unskip\
\newblock
\APACrefYearMonthDay{2026{\protect \BCnt {1}}}{}{}.
\newblock
\APACrefbtitle {{DAC-nipple-predictions} [Data set of model predictions].}
  {{DAC-nipple-predictions} [data set of model predictions].}
\newblock
\APAChowpublished {Hugging Face}.
\newblock
\begin{APACrefURL}
  \url{https://huggingface.co/datasets/h2thez3/DAC-nipple-predictions}
  \end{APACrefURL}
\PrintBackRefs{\CurrentBib}

\bibitem [\protect \citeauthoryear {%
{Hera-MI}%
}{%
{Hera-MI}%
}{%
{\protect \APACyear {2026}}%
{\protect \APACexlab {{\protect \BCnt {2}}}}}]{%
heramiMriAnalysisPaper2026}
\APACinsertmetastar {%
heramiMriAnalysisPaper2026}%
\begin{APACrefauthors}%
{Hera-MI}.%
\end{APACrefauthors}%
\unskip\
\newblock
\APACrefYearMonthDay{2026{\protect \BCnt {2}}}{}{}.
\newblock
\APACrefbtitle {{mri-dce-analysis} [Code repository].} {{mri-dce-analysis}
  [code repository].}
\newblock
\APAChowpublished {GitHub}.
\newblock
\begin{APACrefURL} \url{https://github.com/maystroh/mri-dce-analysis}
  \end{APACrefURL}
\PrintBackRefs{\CurrentBib}

\bibitem [\protect \citeauthoryear {%
{Hera-MI}%
}{%
{Hera-MI}%
}{%
{\protect \APACyear {2026}}%
{\protect \APACexlab {{\protect \BCnt {3}}}}}]{%
sagittalLandmarks2026}
\APACinsertmetastar {%
sagittalLandmarks2026}%
\begin{APACrefauthors}%
{Hera-MI}.%
\end{APACrefauthors}%
\unskip\
\newblock
\APACrefYearMonthDay{2026{\protect \BCnt {3}}}{}{}.
\newblock
\APACrefbtitle {{MRI Sagitall}.} {{MRI Sagitall}.}
\newblock
\APAChowpublished
  {\url{https://huggingface.co/datasets/h2thez3/breast-mri-sagitall-landmarks}}.
\newblock
\APACrefnote{Images redistributed unchanged from TCIA QIN-BREAST under CC BY
  3.0; masks released under the same licence}
\PrintBackRefs{\CurrentBib}

\bibitem [\protect \citeauthoryear {%
Isensee%
, Jaeger%
, Kohl%
, Petersen%
\BCBL {}\ \BBA {} Maier-Hein%
}{%
Isensee%
\ \protect \BOthers {.}}{%
{\protect \APACyear {2021}}%
}]{%
isensee2021nnunet}
\APACinsertmetastar {%
isensee2021nnunet}%
\begin{APACrefauthors}%
Isensee, F.%
, Jaeger, P\BPBI F.%
, Kohl, S\BPBI A\BPBI A.%
, Petersen, J.%
\BCBL {}\ \BBA {} Maier-Hein, K\BPBI H.%
\end{APACrefauthors}%
\unskip\
\newblock
\APACrefYearMonthDay{2021}{}{}.
\newblock
{\BBOQ}\APACrefatitle {nnU-Net: a self-configuring method for deep
  learning-based biomedical image segmentation} {nnu-net: a self-configuring
  method for deep learning-based biomedical image segmentation}.{\BBCQ}
\newblock
\APACjournalVolNumPages{Nature Methods}{18}{2}{203--211}.
\PrintBackRefs{\CurrentBib}

\bibitem [\protect \citeauthoryear {%
Isensee%
\ \protect \BOthers {.}}{%
Isensee%
\ \protect \BOthers {.}}{%
{\protect \APACyear {2025}}%
}]{%
isensee2025nninteractive}
\APACinsertmetastar {%
isensee2025nninteractive}%
\begin{APACrefauthors}%
Isensee, F.%
, Rokuss, M.%
, Kr{\"a}mer, L.%
, Dinkelacker, S.%
, Ravindran, A.%
, Stritzke, F.%
\BDBL {}others%
\end{APACrefauthors}%
\unskip\
\newblock
\APACrefYearMonthDay{2025}{}{}.
\newblock
{\BBOQ}\APACrefatitle {nnInteractive: Redefining 3D promptable segmentation}
  {nninteractive: Redefining 3d promptable segmentation}.{\BBCQ}
\newblock
\APACjournalVolNumPages{arXiv preprint arXiv:2503.08373}{}{}{}.
\PrintBackRefs{\CurrentBib}

\bibitem [\protect \citeauthoryear {%
M{\"u}ller-Franzes%
\ \protect \BOthers {.}}{%
M{\"u}ller-Franzes%
\ \protect \BOthers {.}}{%
{\protect \APACyear {2025}}%
}]{%
mullerfranzes2025odelia}
\APACinsertmetastar {%
mullerfranzes2025odelia}%
\begin{APACrefauthors}%
M{\"u}ller-Franzes, G.%
\BCBT {}\ \BOthersPeriod {.}
\end{APACrefauthors}%
\unskip\
\newblock
\APACrefYearMonthDay{2025}{}{}.
\newblock
{\BBOQ}\APACrefatitle {A {European} multi-center breast cancer {MRI} dataset}
  {A {European} multi-center breast cancer {MRI} dataset}.{\BBCQ}
\newblock
\APACjournalVolNumPages{arXiv preprint arXiv:2506.00474}{}{}{}.
\newblock
\begin{APACrefURL} \url{https://arxiv.org/abs/2506.00474} \end{APACrefURL}
\PrintBackRefs{\CurrentBib}

\bibitem [\protect \citeauthoryear {%
Rokuss%
, Hamm%
, Kirchhoff%
\BCBL {}\ \BBA {} Maier-Hein%
}{%
Rokuss%
\ \protect \BOthers {.}}{%
{\protect \APACyear {2025}}%
}]{%
rokuss2025dac}
\APACinsertmetastar {%
rokuss2025dac}%
\begin{APACrefauthors}%
Rokuss, M.%
, Hamm, B.%
, Kirchhoff, Y.%
\BCBL {}\ \BBA {} Maier-Hein, K.%
\end{APACrefauthors}%
\unskip\
\newblock
\APACrefYearMonthDay{2025}{}{}.
\newblock
{\BBOQ}\APACrefatitle {Divide and Conquer: A Large-Scale Dataset and Model for
  Left--Right Breast MRI Segmentation} {Divide and conquer: A large-scale
  dataset and model for left--right breast mri segmentation}.{\BBCQ}
\newblock
\APACjournalVolNumPages{arXiv preprint arXiv:2507.13830}{}{}{}.
\newblock
\begin{APACrefURL} \url{https://arxiv.org/abs/2507.13830} \end{APACrefURL}
\newblock
\APACrefnote{MICCAI 2025 WOMEN workshop. Dataset and model:
  \url{https://github.com/MIC-DKFZ/BreastDivider}; dataset V2:
  \url{https://huggingface.co/datasets/Bubenpo/BreastDividerDataset}}
\PrintBackRefs{\CurrentBib}

\bibitem [\protect \citeauthoryear {%
Saha%
\ \protect \BOthers {.}}{%
Saha%
\ \protect \BOthers {.}}{%
{\protect \APACyear {2021}}%
}]{%
saha2021duke}
\APACinsertmetastar {%
saha2021duke}%
\begin{APACrefauthors}%
Saha, A.%
, Harowicz, M\BPBI R.%
, Grimm, L\BPBI J.%
, Weng, J.%
, Cain, E\BPBI H.%
, Kim, C\BPBI E.%
\BDBL {}Mazurowski, M\BPBI A.%
\end{APACrefauthors}%
\unskip\
\newblock
\APACrefYearMonthDay{2021}{}{}.
\newblock
\APACrefbtitle {Dynamic contrast-enhanced magnetic resonance images of breast
  cancer patients with tumor locations [data set].} {Dynamic contrast-enhanced
  magnetic resonance images of breast cancer patients with tumor locations
  [data set].}
\newblock
\APAChowpublished {The Cancer Imaging Archive}.
\PrintBackRefs{\CurrentBib}

\bibitem [\protect \citeauthoryear {%
Wald%
\ \protect \BOthers {.}}{%
Wald%
\ \protect \BOthers {.}}{%
{\protect \APACyear {2025}}%
}]{%
wald2025crosstechnique}
\APACinsertmetastar {%
wald2025crosstechnique}%
\begin{APACrefauthors}%
Wald, T.%
, Hamm, B.%
, Holzschuh, J\BPBI C.%
, El~Shafie, R.%
, Kudak, A.%
, Kovacs, B.%
\BDBL {}others%
\end{APACrefauthors}%
\unskip\
\newblock
\APACrefYearMonthDay{2025}{}{}.
\newblock
{\BBOQ}\APACrefatitle {Enhancing deep learning methods for brain metastasis
  detection through cross-technique annotations on SPACE MRI} {Enhancing deep
  learning methods for brain metastasis detection through cross-technique
  annotations on space mri}.{\BBCQ}
\newblock
\APACjournalVolNumPages{European Radiology Experimental}{9}{1}{1--14}.
\PrintBackRefs{\CurrentBib}

\end{thebibliography}

\onecolumn
\appendix
\setcounter{table}{0}
\renewcommand{\thetable}{A\arabic{table}}
\renewcommand{\theHtable}{A\arabic{table}}

\section{Evaluation on unseen sagittal data}
\label{sec:sagittal}

We evaluated the models on 65 sagittal cases excluded from training:
30 from MAMA-MIA (26 I-SPY1 and 4 NACT cases), with annotations that are not publicly released,
and 35 from the publicly available QIN-BREAST-derived landmark dataset
\citepalias{sagittalLandmarks2026}. The public dataset contains unilateral
pre-contrast MRI volumes, breast-region and nipple masks, and case metadata;
its dataset card provides the annotation and acquisition details.
Table~\ref{tab:sagittal} reports nipple true-positive rate (TPR), missed
references and false-positive (FP) cases. TPR is the proportion of cases with
a detected reference nipple; an FP case contains at least one spurious
predicted nipple component. The released per-breast model detects a reference
nipple in 26 of 30 MAMA-MIA cases (86.7\% TPR) and 32 of 35 QIN cases
(91.4\% TPR). Its predictions include spurious components in 3 MAMA-MIA
cases and 4 QIN cases. These results show partial transfer to unseen sagittal
acquisitions, with missed nipples and false-positive predictions remaining.

\begin{table}[H]
\centering
\caption{Evaluation on unseen sagittal acquisitions: MAMA-MIA subset ($n=30$)
and QIN ($n=35$). TPR is the proportion of cases with a detected reference
nipple. Missed is the number of cases without a predicted reference nipple;
FP cases contain at least one spurious predicted nipple component.
Mean distance is the mean Euclidean distance between the centroids of the
largest annotated and predicted nipple components, computed over cases
with a predicted nipple.}
\label{tab:sagittal}
\begin{tabular}{llrrrr}
\toprule
Model & Cohort & TPR & Missed & FP cases & Mean distance (mm) \\
\midrule
Compact, per-breast (released) & MAMA-MIA & 86.7\% & 4 & 3 & 4.49 \\
nnU-Net, whole volume & MAMA-MIA & 96.7\% & 1 & 21 & 6.21 \\
Compact, whole volume & MAMA-MIA & 93.3\% & 2 & 12 & 5.33 \\
\midrule
Compact, per-breast (released) & QIN & 91.4\% & 3 & 4 & 7.82 \\
nnU-Net, whole volume & QIN & 100.0\% & 0 & 34 & 10.85 \\
Compact, whole volume & QIN & 91.4\% & 3 & 18 & 6.23 \\
\bottomrule
\end{tabular}
\end{table}

\subsection{Cases with missed nipple detections}

\begin{figure}[!htbp]
\centering
\includegraphics[width=0.49\linewidth]{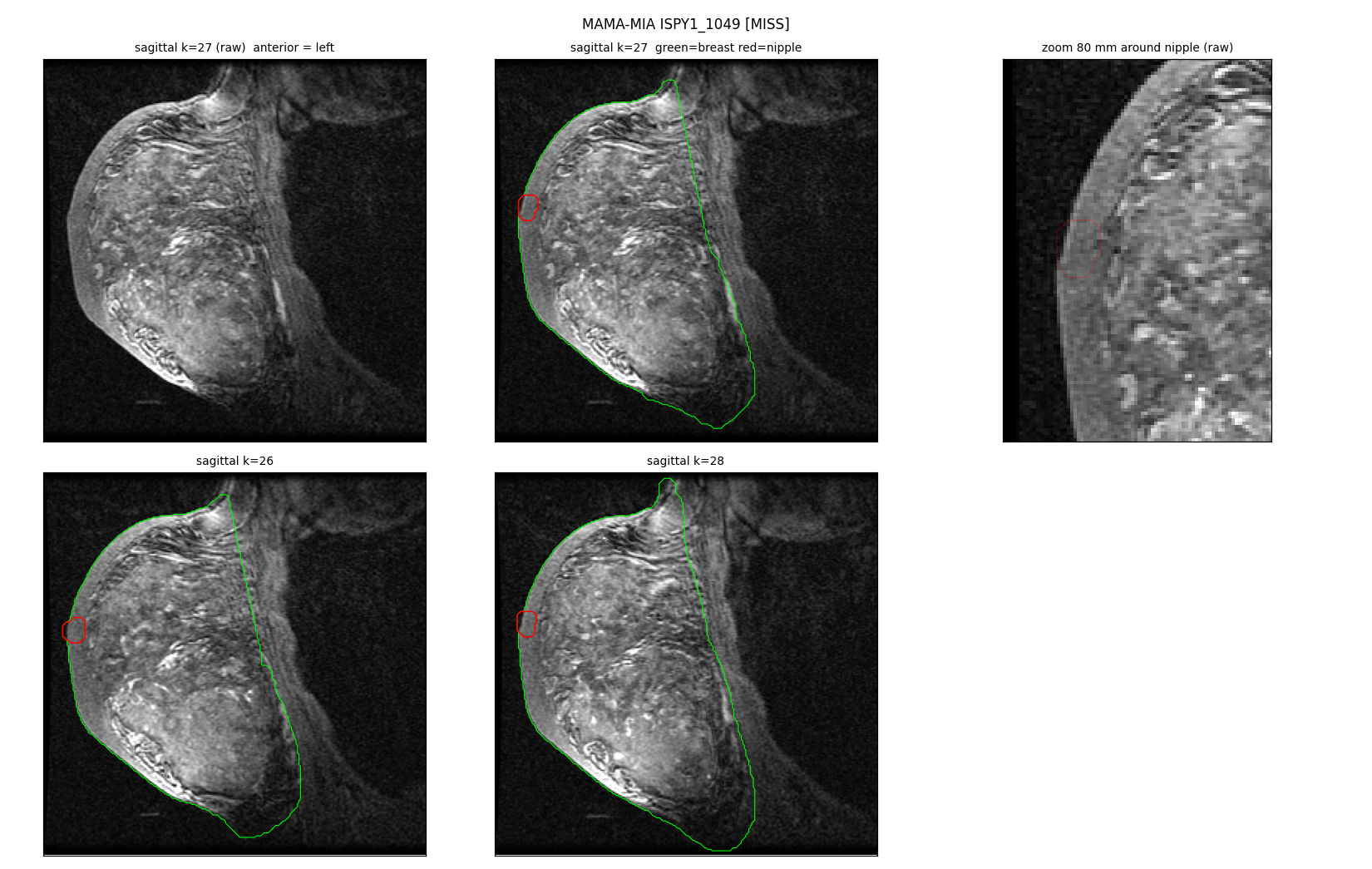}\hfill
\includegraphics[width=0.49\linewidth]{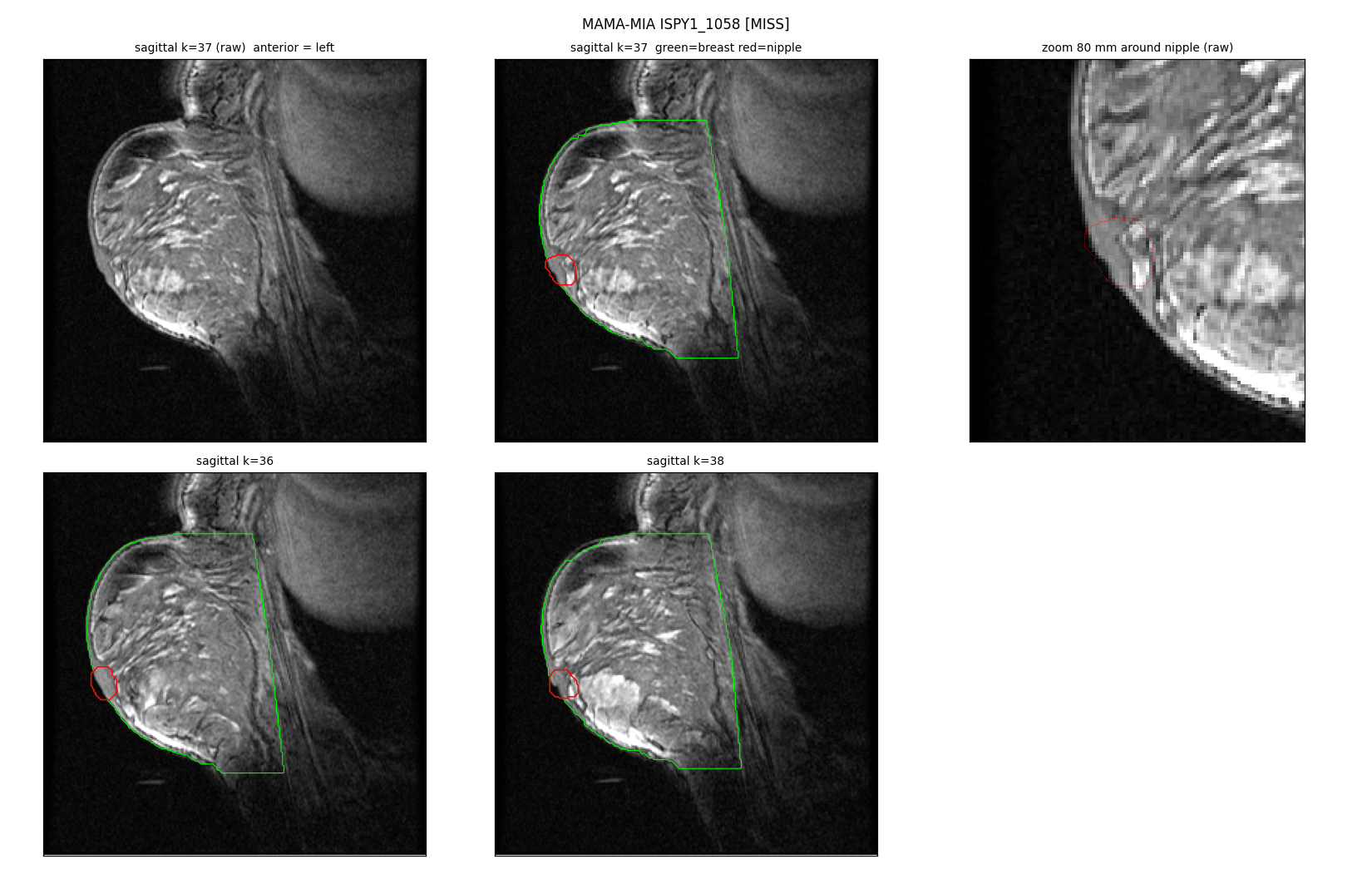}
\par\medskip
\includegraphics[width=0.49\linewidth]{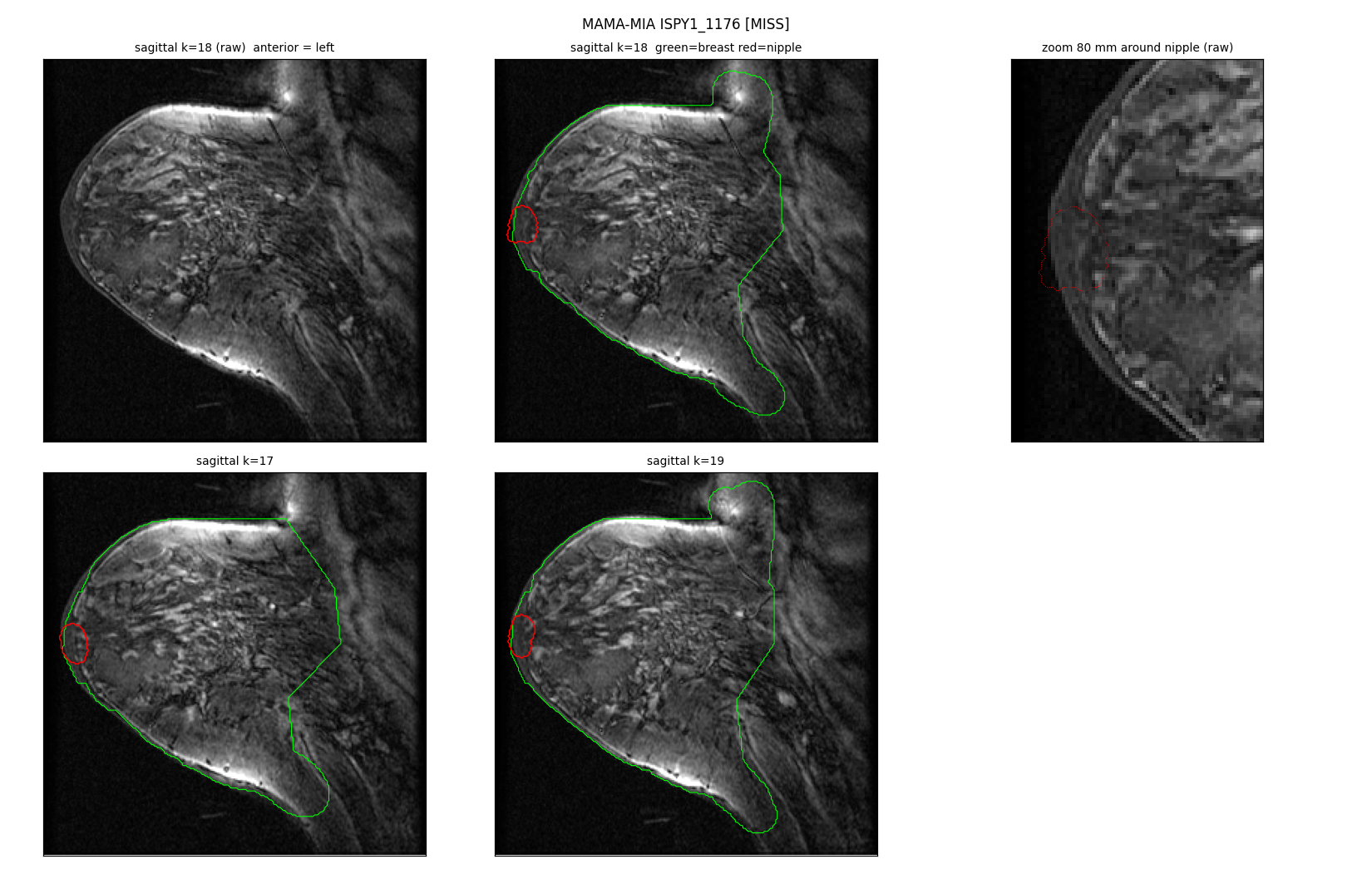}\hfill
\includegraphics[width=0.49\linewidth]{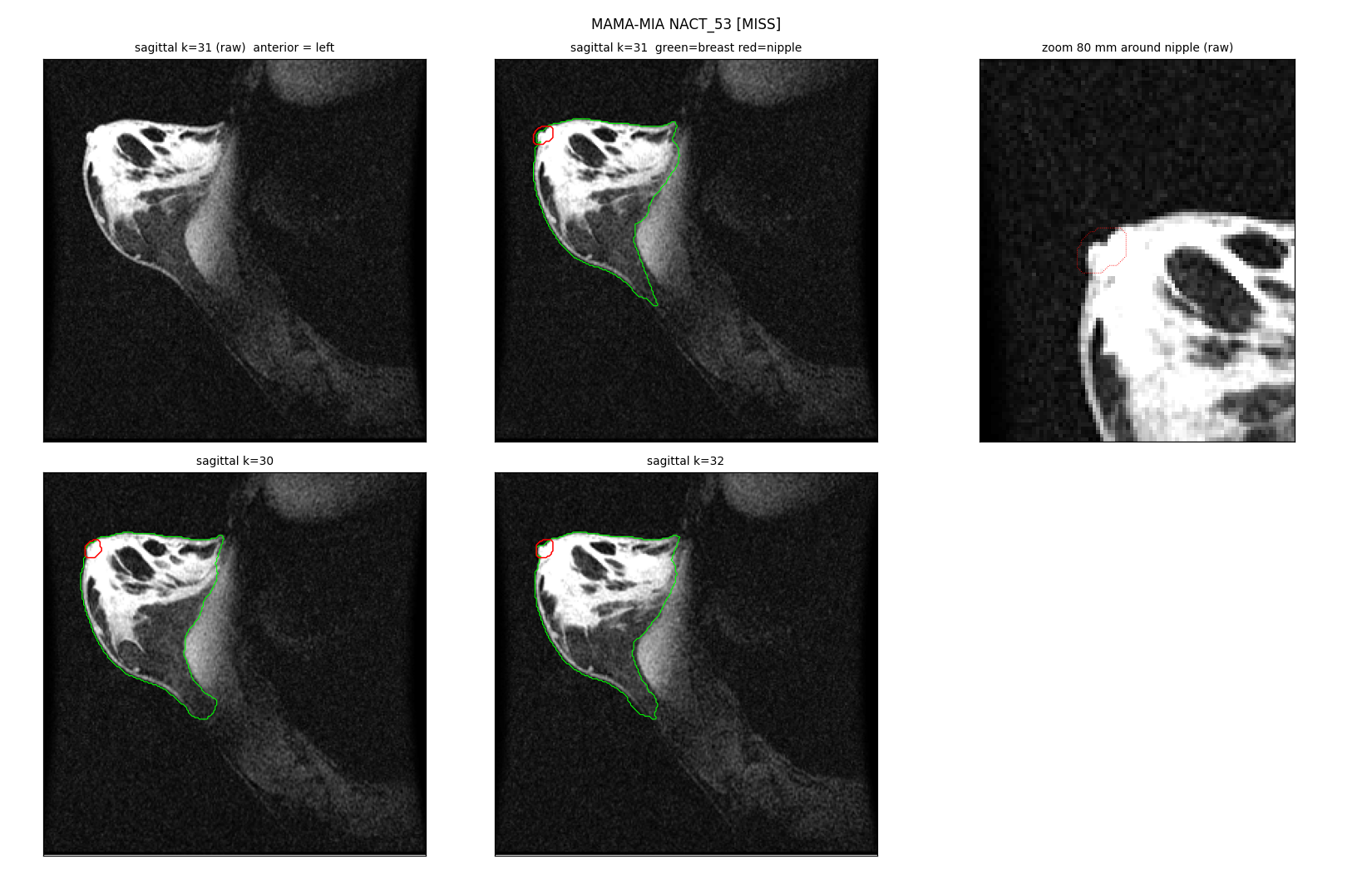}
\caption{MAMA-MIA cases with no predicted nipple. From top left to bottom
right: \texttt{ISPY1\_1049}, \texttt{ISPY1\_1058}, \texttt{ISPY1\_1176}
and \texttt{NACT\_53}. Green contours mark the reference breast region and
red contours mark the reference nipple where it appears in the slice.}
\label{fig:sagittal-misses-mamamia}
\end{figure}

\begin{figure}[!htbp]
\centering
\includegraphics[width=0.49\linewidth]{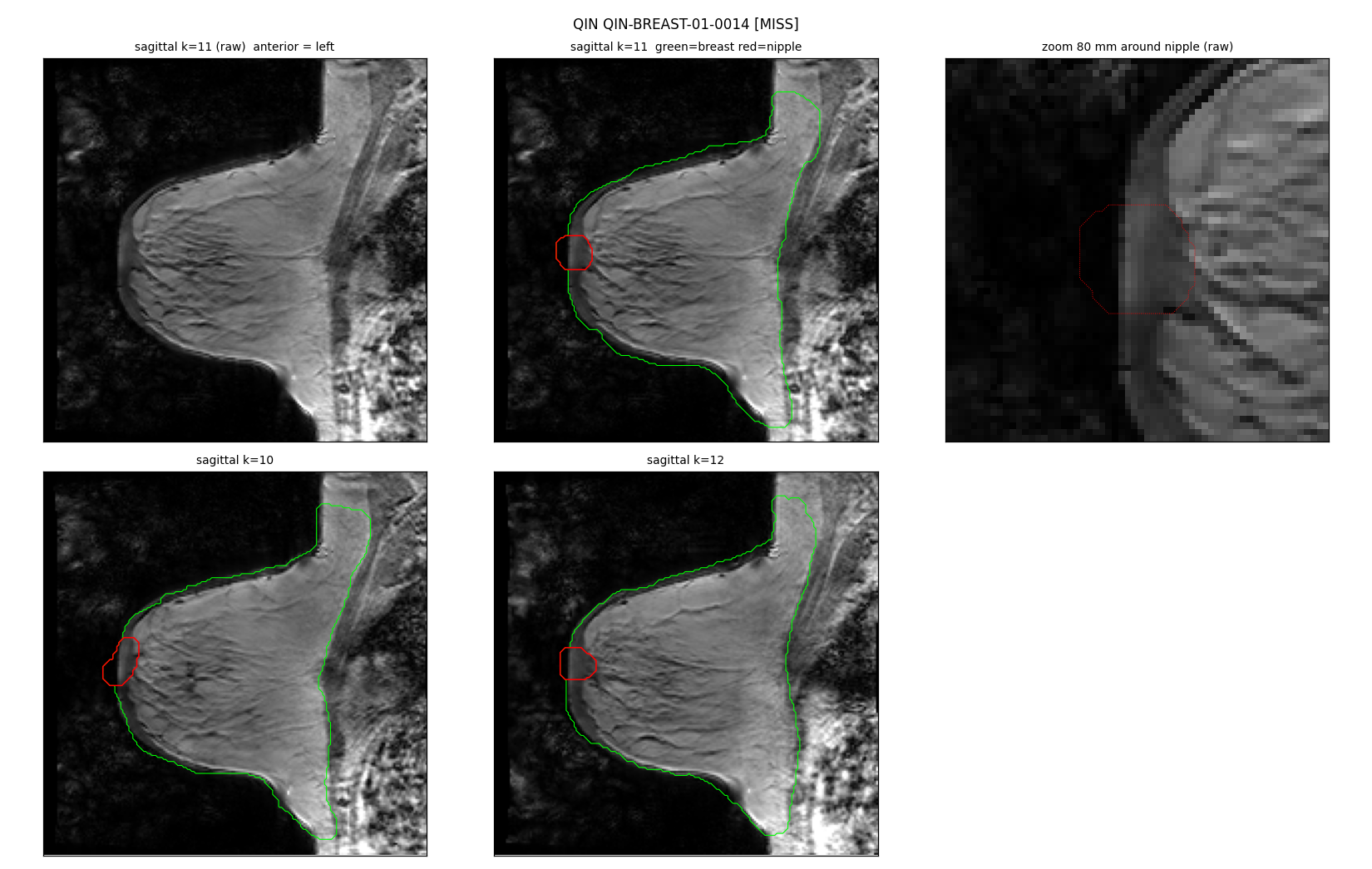}\hfill
\includegraphics[width=0.49\linewidth]{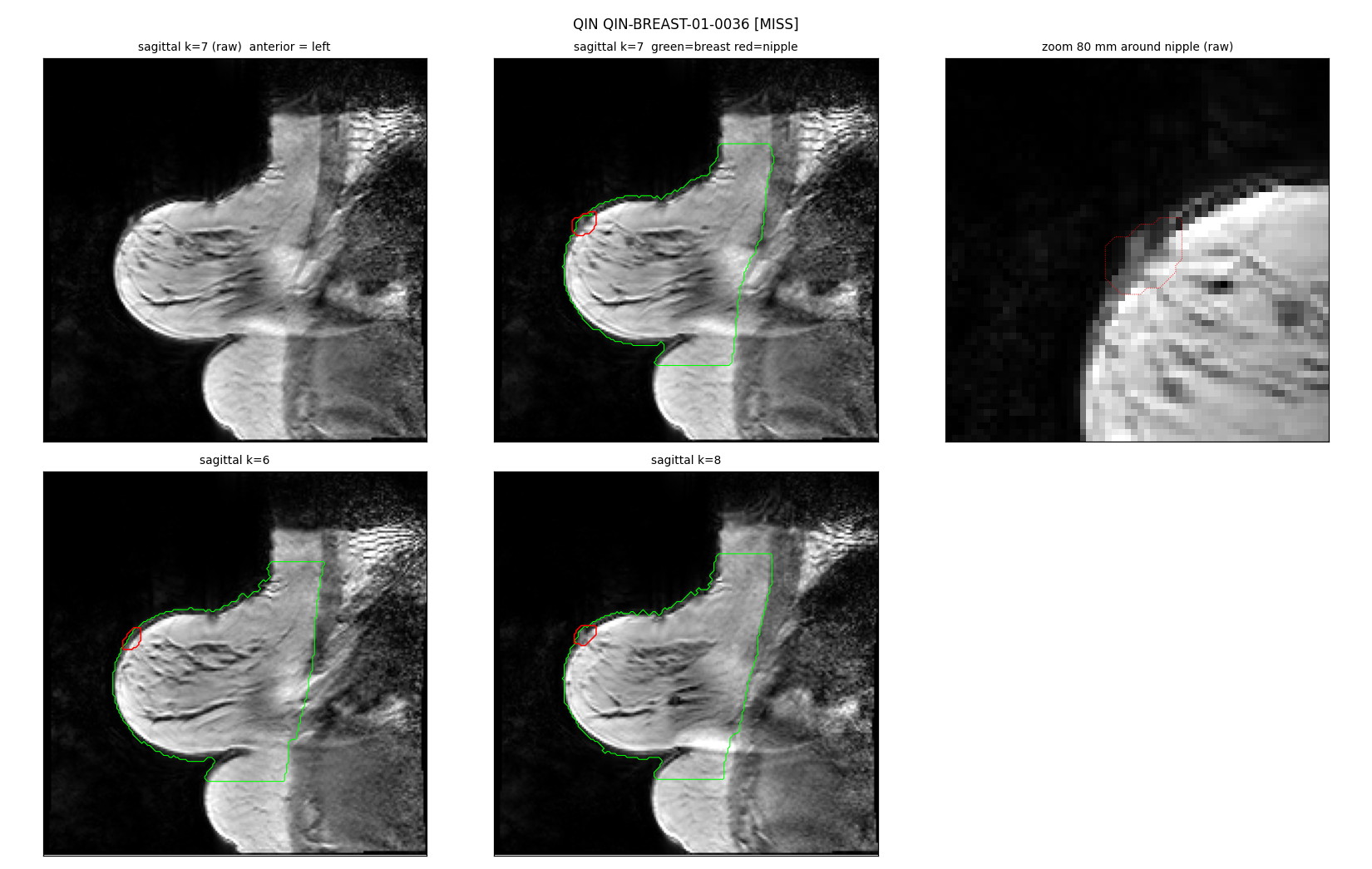}
\par\medskip
\includegraphics[width=0.49\linewidth]{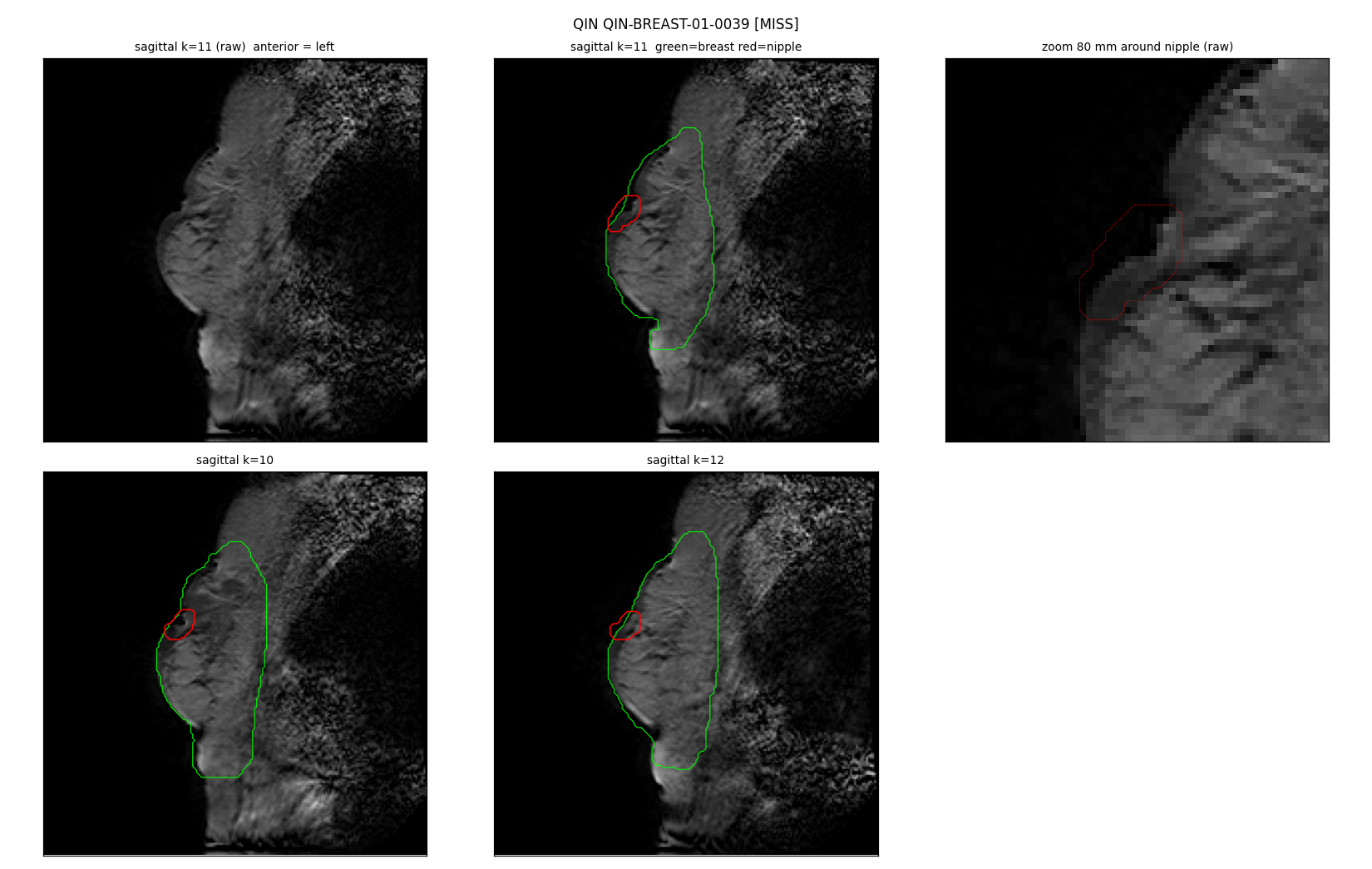}
\caption{QIN cases with no predicted nipple. From top left to bottom:
\texttt{QIN-BREAST-01-0014}, \texttt{QIN-BREAST-01-0036} and
\texttt{QIN-BREAST-01-0039}. The reference contours use the same colours as
Figure~\ref{fig:sagittal-misses-mamamia}.}
\label{fig:sagittal-misses-qin}
\end{figure}

The released compact per-breast model produced no nipple prediction in
seven sagittal cases: four from MAMA-MIA (Figure~\ref{fig:sagittal-misses-mamamia})
and three from QIN (Figure~\ref{fig:sagittal-misses-qin}).

\clearpage
\end{document}